\documentclass[10pt,twocolumn,letterpaper]{article}

\usepackage{cvpr}              

\usepackage{booktabs}
\usepackage{lipsum}
\usepackage{multirow}
\usepackage{makecell}
\usepackage{adjustbox}
\usepackage[dvipsnames]{xcolor}

\definecolor{cvprblue}{rgb}{0.21,0.49,0.74}
\usepackage[pagebackref,breaklinks,colorlinks,citecolor=cvprblue]{hyperref}

\def\paperID{277} 
\def\confName{3DV\xspace}
\def\confYear{2024\xspace}

\title{NeVRF: Neural Video-based Radiance Fields for Long-duration Sequences}

\author{{ Minye Wu  
\qquad 
Tinne Tuytelaars} \\
{ KU Leuven}\\
{\tt\small \{minye.wu, tinne.tuytelaars\}@esat.kuleuven.be}
}

\begin{document}
\maketitle
\begin{abstract}
Adopting Neural Radiance Fields (NeRF)  to long-duration dynamic sequences has been challenging.
Existing methods struggle to balance between quality and storage size and encounter difficulties with complex scene changes such as topological changes and large motions. 
To tackle these issues, we propose a novel neural video-based radiance fields (NeVRF) representation. 
NeVRF marries neural radiance field with image-based rendering to support photo-realistic novel view synthesis on long-duration dynamic inward-looking scenes. 
We introduce a novel multi-view radiance blending approach to predict radiance directly from multi-view videos.
By incorporating continual learning techniques, NeVRF can efficiently reconstruct frames from sequential data without revisiting previous frames, enabling long-duration free-viewpoint video. 
Furthermore, with a tailored compression approach, NeVRF can compactly represent dynamic scenes, making dynamic radiance fields more practical in real-world scenarios. 
Our extensive experiments demonstrate the effectiveness of NeVRF in enabling long-duration sequence rendering, sequential data reconstruction, and compact data storage.
\end{abstract}

\section{Introduction}

Neural Radiance Field (NeRF)~\cite{mildenhall2021nerf} has facilitated a series of breakthroughs in novel view synthesis, enriching the contents for virtual reality, telecommunications, etc. 
%
Among them is photo-realistic free-viewpoint video~(FVV). 
%
However, generating dynamic radiance fields in practical settings remains challenging due to their storage requirements and the complexity of processing streaming input data.

Recent advances have improved NeRF content generation in many ways, such as accelerating the training speed~\cite{deng2022depth, sun2022direct}, rendering speed~\cite{reiser2021kilonerf, garbin2021fastnerf, neff2021donerf}, and reducing the storage in a compact representation~\cite{chen2022tensorf, takikawa2022variable}. 
These methods mainly focus on static scenes. 
Some methods introduce space deformation~\cite{pumarola2021d,liu2022devrf,park2021nerfies} to handle dynamic scenes.
They disentangle motion and canonical space from sequences to achieve fast training speed with sparse views and further speed up by leveraging an explicit grid representation~\cite{liu2022devrf}. 
%
%
However, the decreased number of views can impair performance in scenes with large motion, and limits their ability to handle topological changes due to the use of canonical space.
%
%
MLPs~\cite{pumarola2021d,zhang2021editable} and space-time latent codes~\cite{li2022neural} are used to capture motion information. 
While they achieve relatively compact storage, they have large computational complexity and their performance is constrained by the network capacity, which makes it challenging to deal with long-duration or large motion sequences.
In another line of work, grid-based methods usually require a large memory~\cite{liu2022devrf}, making the transmission and storage of NeRF content challenging. 
%
%
%
%

\begin{figure}[t]
	\begin{center}
		\includegraphics[width=1.0\linewidth]{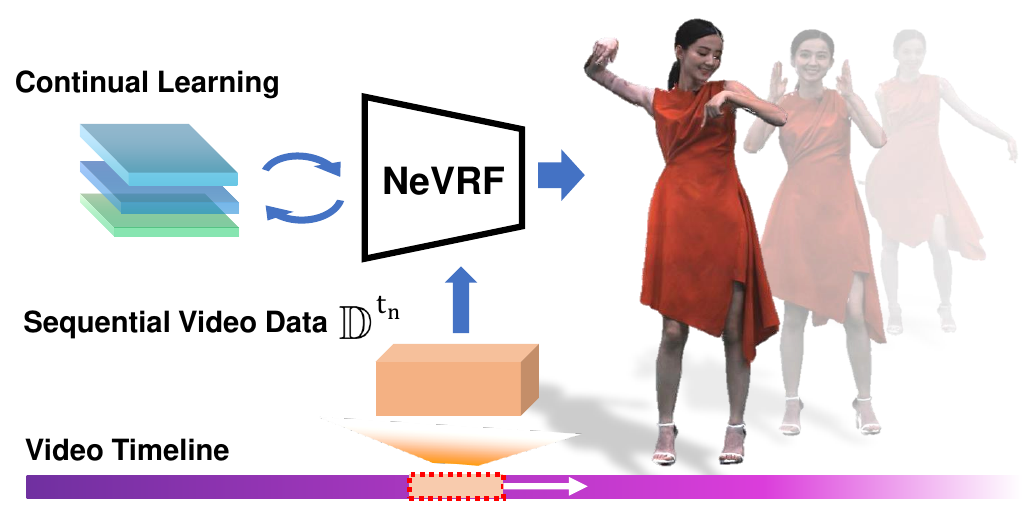}
	\end{center}
	\vspace{-0.7cm}
	\caption{Our method introduces a neural video-based radiance field representation for dynamic scenes achieving photo-realistic novel view synthesis with small storage (around 1.32MB per-frame) in sequential input setting. NeVRF can continuously reconstruct a sequence without revisiting previous data, making it suitable for long-duration sequences.}
	\label{fig:teaser}
	\vspace{-5mm}
\end{figure}

%
Furthermore, previous approaches require all sequence data to be available throughout the entire training process to enable i.i.d. sampling of rays for assembling training batches. 
%
This becomes unfeasible for long-duration or endless sequences owing to the substantial surge in memory consumption and the indeterminate length of data. 
%
%
Dynamic scene data is naturally sequential and ordered by time, but 
so far this property has not been exploited for more efficient FVV generation.
%
%
%
%
%

In this paper, we present a novel Neural Video-based Radiance Field~(NeVRF) representation to tackle the issues and challenges of long-duration sequences with sequential input, as illustrated in Figure~\ref{fig:teaser}.  
%
%
NeVRF directly infers the radiance fields from the inward-looking input videos frame-per-frame.
At the core is a Multi-view Radiance Blending approach that predicts colors from multi-view frame images. 
Specifically, we deploy a shared lightweight feature encoder to extract local context and semantic cues from multi-view frame images. 
%
Another network learns to predict views' visibility and blending weights for sample points along rays, avoiding traditional high-cost visibility calculations. 
%
%
The density fields are represented using explicit volumetric grids, enabling it to be decoupled from the appearance rendering process. 
This separative design enables both high-quality rendering and efficient storage compression.
Our representation leverages off-the-shelf video codecs and our proposed density fields compression algorithm to achieve compact storage. 
%

%
We also introduce a tailored continual learning paradigm for the setting. 
We leverage replay-based continual learning~\cite{rebuffi2017icarl,rolnick2019experience} against catastrophic forgetting~\cite{mccloskey1989catastrophic, de2021continual,isele2018selective} of the old network knowledge. 
NeVRF caches important rays instead of images in the experience buffer, avoiding the requirement of accessing previous data during training. 
This inter-frame independent design makes NeVRF well-suited for long sequences while maintaining a fixed memory footprint regardless to the length of the sequences.
The introduction of continual learning allows NeVRF to reconstruct new frames while still performing well on previous frames. 
So the proposed method can pause and resume learning at any point and playback the learned previous frames instantly.
%
%
The rendering pipeline is differentiable so that networks and density volumes can be jointly optimized in an end-to-end fashion. 

In summary, our contributions are:
\begin{itemize}
  \item We introduce a novel video-based radiance field representation with separative geometry and appearance information to achieve compact storage for dynamic sequences. 
  \item We propose a novel image-based rendering pipeline paired with the representation, achieving high-quality results.
  \item We present a new continual learning paradigm to deal with sequential data, supporting long-duration or endless sequences.
\end{itemize}
\section{Related Work}

\vspace{1mm}\noindent{\bf Image-based Rendering.} 
Early work on IBR synthesizes novel view images by blending color pixels from a set of reference views. 
Most of IBR methods require an explicit proxy geometry, like a plane~\cite{levoy1996light}, 3D meshes~\cite{debevec1996modeling, gortler1996lumigraph, buehler2001unstructured}, layered depth~\cite{shade1998layered,daribo2011novel}, or silhouettes~\cite{matusik2000image,Matusik2002Image}, to render images at new viewpoints. 
The rendering quality of these methods depends on the camera setting~\cite{levoy1996light,Matusik2002Image} and the accuracy of the geometry~\cite{buehler2001unstructured,gortler1996lumigraph}. 
Defects in geometry, such as missing or inaccuracy parts, will cause artifacts, e.g. ghosting effects. 
Recent work alleviates this problem by leveraging additional information or knowledge, such as optical flow~\cite{casas20154d,du2018montage4d} and soft blending~\cite{penner2017soft,riegler2020free}. 
Nevertheless, they are still fundamentally limited by the geometry accuracy. 
Conventional 3D reconstruction methods~\cite{campbell2008using, tola2012efficient} rely on multi-view input images' quality. 
Textureless or reflective regions on images will cause failures easily, and these methods cannot handle semi-transparent surfaces. 
In contrast to the traditional image-based rendering methods, we use density fields as the proxy geometry and build an end-to-end differentiable rendering pipeline. 
As a result we can take advantage of neural rendering to avoid some issues of 3D reconstruction. 

\begin{figure*}[t]
	\begin{center}
		\includegraphics[width=0.9\linewidth]{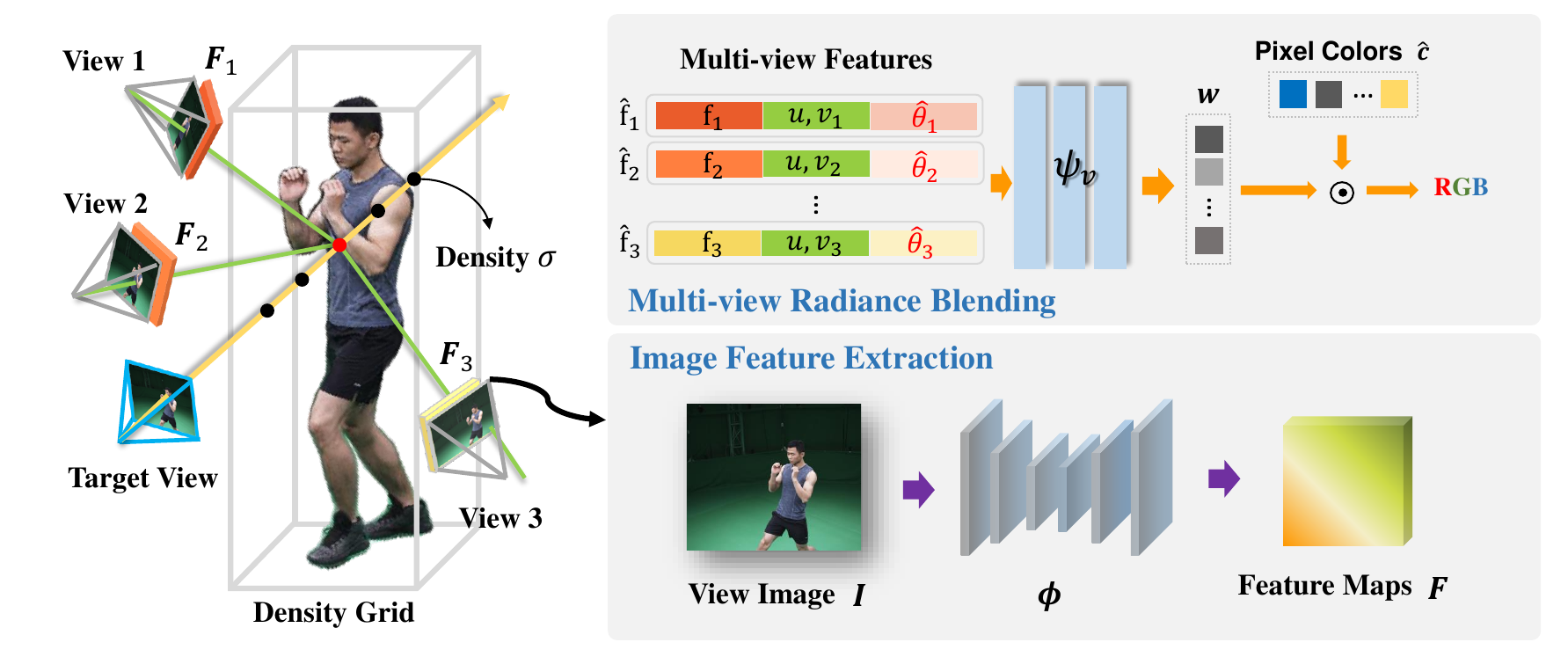}
	\end{center}
	\vspace{-0.5cm}
	\caption{Illustration of the proposed rendering pipeline under the novel neural video-based radiance field representation. NeVRF extracts feature maps from selected reference views. 
 $\mathbf{f}_i$ are the extracted multi-view features, and $u$ is their mean; $v_i$ and $\hat{\theta}_i$ are feature variances and relative viewing directions respectively. 
 NeVRF exploits the multi-view radiance blending method to predict its RGB color as the density is directly interpolated on the density grid.}
	\vspace{-3mm}
	\label{fig:mrb}

\end{figure*}

\vspace{1mm}\noindent{\bf Neural Representations for Novel View Synthesis.}  
The recent progress of neural rendering shows the ability to achieve photo-realistic novel view synthesis. 
They are based on various scene representations, such as point clouds~\cite{aliev2020neural, wu2020multi}, voxels~\cite{liu2019neural} and implicit functions~\cite{mescheder2019occupancy,saito2019pifu}. 
The emergence of Neural Radiance Fields (NeRF)~\cite{mildenhall2021nerf} has greatly boosted the development of scene modeling and rendering techniques. 
NeRF exploits coordinate-based networks to represent radiance fields and introduces an end-to-end differentiable rendering pipeline with volume rendering. 
DVGO~\cite{sun2022direct} replaces MLPs in NeRF with grid-based representations, and Instant-NGP~\cite{muller2022instant} leverages a multi-resolution hash table to represent radiance fields. 
Other methods~\cite{liu2020neural, yu2021plenoctrees} utilize special volumetric data structures to access scene space efficiently. 
They have greatly improved the training and rendering efficiency but require a considerable amount of memory. 
IBRnet~\cite{wang2021ibrnet} infers the scene purely from multi-view images and has generalization ability to new scenes, but it has  difficulty with domain gaps. 
Different from IBRnet, our method represents geometry using explicit density grids instead of predicting directly density from reference views, enabling high-accurate rendering.
Neural radiance fields can also be represented in tensors~\cite{chen2022tensorf}, which has a fast training and rendering speed while having a relatively compact storage. 
However, these methods mainly focus on static scenes while dynamic scenes have an exponentially larger amount of data.

\vspace{1mm}\noindent{\bf Neural Rendering for Dynamic Scenes.} 
Rendering dynamic scenes with neural rendering is an important branch of novel view synthesis. 
Previous work exploits multi-view image features~\cite{lombardi2019neural}, a coarse proxy geometry~\cite{wu2020multi,peng2021neural} paired with a differentiable rendering pipeline to handle dynamic scenes. 
With the emergence of neural radiance field techniques, how to encode temporal information in an implicit representation becomes the key problem. 
Instead of inquiring 3D points in a radiance field representation, learnable latent codes for individual frame~\cite{li2022neural}, spatio-temporal coordinates~\cite{xian2021space,du2021neural}, and Fourier coefficients~\cite{wang2022fourier} are used to model dynamic sequences. 
However, the length of sequences is limited due to the network capacity, and their fully implicit representations cause high computational complexity. 
Another stream of dynamic NeRF methods~\cite{pumarola2021d,zhang2021editable,liu2022devrf,park2021nerfies} disentangle motions from the scene as a deformation field and predict colors and densities by mapping point coordinates into a canonical space. 
They can model dynamic scenes efficiently and support sparse inputs, but they tend to fail when the scene has large motion and topological changes due to the use of the stationary canonical space. 
A series of grid-based methods~\cite{liu2022devrf, fang2022fast} dramatically accelerates the training and rendering speed but bring high storage cost~\cite{liu2022devrf}. 
The recent method, ENeRF~\cite{lin2022efficient}, exploits neural networks to inference per-frame geometry via multi-view consistency and achieves efficient rendering. 
\cite{zhao2022human} leverages an explicit animated mesh with multi-view neural blending scheme~\cite{suo2021neuralhumanfvv} to balance quality and bandwidth, but it can only handle scenes without translucent parts. 
Our method aims at dealing with long-duration dynamic scenes, with low storage cost and fine details.

\section{Overview}

The proposed NeVRF marries neural radiance fields with image-based rendering and continual learning techniques to support free-viewpoint photo-realistic view synthesis on long duration dynamic scenes. 
%
Assuming that sequential data $\mathbb{D}_{t_1}^{t_2}=\{\mathcal{D}^{t_1}, ..., \mathcal{D}^{t_2}\}$ are fed into our algorithm one by one. $\mathcal{D}^t=\{\mathbf{I}^t_i, \mathbf{P}^t_i\}_{i=1}^{N_c}$ contains multi-view images $\mathbf{I}^t_i$ at time instance $t$ and camera projection matrices $\mathbf{P}^t_i$ of the corresponding views, where $N_c$ is the number of views. 
%
%

We design a three-step pipeline to process every incoming sequential data. 
%
In the first step, we leverage a fast reconstruction approach to acquire coarse geometry. 
%
In the second step, we introduce a multi-view based rendering method on the density field and pair it with a tailored learning scheme to adapt the model to the newest frames and previous frames to maintain high fidelity rendering quality. 
In the third step, we propose a novel compression solution to achieve small storage size. 

\section{Neural Video-based Radiance Field}

In this section, we first review the Neural Radiance Fields in~\S~\ref{sec:pre}. Then we propose a novel Multi-view Radiance Blending method to take advantage of compact video representations (\S~\ref{sec:mrb}). Lastly, we introduce the three-step pipeline for sequential data with continual learning to enable long-duration sequence rendering and lightweight storage (\S~\ref{sec:cl}).

\subsection{Preliminaries}\label{sec:pre}
%
NeRF~\cite{mildenhall2021nerf} models a 3D scene with a function $\Psi$ which takes as input the coordinate $\mathbf{x}\in\mathbb{R}^3$ and view direction $\mathbf{d} = (\theta,\phi)$ for each point in the space and maps them into a color $\mathbf{c}$ and density $\sigma$ to represent the properties of the point: 
\begin{equation}
\begin{split}
\mathbf{c}, \sigma &= \Psi(\mathbf{x}, \mathbf{d}). 
\end{split}
\label{eq:NeRF}
\end{equation}
For each pixel in the target view, NeRF exploits volume rendering equations to accumulate the properties of sample points along camera rays and obtain their pixel colors. 
%
%
%
%
DVGO~\cite{sun2022direct} follows a similar rendering pipeline and leverages explicit and discrete grid-based representations to achieve fast training and rendering speed. 
In our proposed method, inspired by DVGO, the scene geometry is represented in the form of density grids $\mathbf{V}^t_{\sigma}$.
The raw density of sample point $\mathbf{x}$ is obtained by:
\begin{equation}
\begin{split}
\Ddot{\sigma} &=  interp(\mathbf{x},\mathbf{V}^t_{\sigma}) \\
\end{split}
\label{eq:density}
\end{equation}
where $interp(\cdot)$ is a trilinear interpolation function on the grids.
%
In terms of color prediction, we utilize multi-view videos and propose a Multi-view Radiance Blending scheme to avoid the large data size of grids with high-dimensional features.
%

%
%
%

\subsection{Multi-view Radiance Blending}\label{sec:mrb} 
Our proposed Multi-view Radiance Blending aims at predicting colors $\mathcal{C}=\{\mathbf{c}_i\}$ for sample points $\mathcal{X}=\{\mathbf{x}_i\}$ using images from multi-view data $\mathcal{D}^t$. 
%
%
Specifically, for each sample point, we first select $k$ nearby camera views from $\mathcal{D}^t$ by selecting the top k views with smallest angular difference, denoted as $\mathcal{\hat{D}}^t=\{\mathbf{I}^t_j,\mathbf{P}^t_j\}_{j=1}^{k}$.
%
%
%
%
Simply blending pixel colors from all viewpoints, as done in~\cite{buehler2001unstructured}, can cause severe ghosting artifacts, especially when including viewpoints where the point is not visible. Alternatively, computing visibility for source views on the density representation is costly; it requires depth values by sampling additional points.
%

We set out to aggregate features from multi-view data and determine their visibility directly inside a forward pass.  
We use a pre-trained convolutional neural network to extract a feature map $\mathbf{F}_j^t$ from each image $\mathbf{I}^t_j$. 
For each sample point $\mathbf{x}_i$, we back-project it to all source views in $\mathcal{\hat{D}}^t$ and then collect features $\{\mathbf{f}_{i,j}^t\}_{j=1}^k$ and pixel colors $\{\mathbf{\hat{c}}_{i,j}^t\}_{j=1}^k$. 
%
%
Then we deploy Multilayer Perceptrons~(MLPs) to infer the visibility weights. 
More specifically, the proposed pipeline is illustrated in Figure~\ref{fig:mrb}. 
The blending network $\Psi_{v}$ aims at predicting weights for selected views. 
Views that cannot see the sample point will be assigned weights very close to zero. 
%
%
We compute the mean $\mathbf{u}$ and per-view variance $\mathbf{v}_j$ from features vectors $\{\mathbf{f}_{i,j}^t\}_{j=1}^k$ to capture global information. 
%
To account for geometric relationships, we incorporate relative viewing direction information $\{\mathbf{\hat{\theta}}_{i,j}\}$, which are calculated as the cosine distance between the vectors of target ray and the pixel rays where the point is back-projected onto the selected view.
We concatenate them along with original features vectors to form a new feature vector $\mathbf{\hat{f}}_{i,j}^t=[\mathbf{f}_{i,j}^t,\mathbf{\hat{\theta}}_{i,j},\mathbf{v}_j,\mathbf{u}]$. 
$\Psi_{v}$ determines weights $\mathbf{w}_i^t\in\mathbb{R}^k$ of views based on local feature and semantics multi-view consistency of $\{\mathbf{\hat{f}}_{i,j}^t\}_{j=1}^k$, which is formulated as: 
\begin{equation}
\begin{split}
\mathbf{w}_i^t &= \Psi_{v}(\{\mathbf{\hat{f}}_{i,j}^t\}_{j=1}^k), 
\end{split}
\label{eq:MLP1}
\end{equation}
where $\sum\mathbf{w}_i^t=1.0$.
%
%
%
The Blended point color is formulated as: 
\begin{equation}
\begin{split}
\mathbf{c}_i^t & = \sum_j w_{i,j}^t\cdot\mathbf{\hat{c}}_{i,j}^t, 
\end{split}
\label{eq:MLP2}
\end{equation}
%
Combining the Equ.~\ref{eq:MLP1}, \ref{eq:MLP2}, and \ref{eq:density}, we can obtain the properties of sample points and render novel views via volume rendering.
%
Note that this is a differentiable rendering pipeline. 
Therefore, we are able to optimize both the network parameters and the density grid during training. 
Our method separates density and color inference, avoiding the need for visibility reasoning along the entire rays in IBRnet~\cite{wang2021ibrnet}, leading to faster inference speeds and more accurate, deterministic geometry modeling.

\subsection{Sequential Learning Scheme}\label{sec:cl}
We also present a novel learning scheme tailored for sequential data to reconstruct geometry and update the networks efficiently. 
%
%
We process each newly coming data $\mathbb{D}_{t_1}^{t_2}$ with the following three training steps in order. 
%

\begin{figure*}[t]
	\begin{center}
		\includegraphics[width=0.8\linewidth]{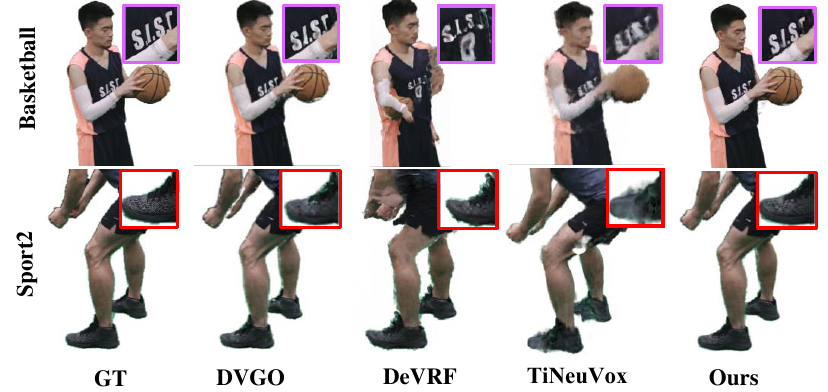}
	\end{center}
	\vspace{-5mm}
	\caption{Qualitative comparisons on 'Basketball' and 'Sport2' scenes in NHR~\cite{wu2020multi} dataset. These are rendered test images from $50$-th frame of the sequences.}
	\label{fig:comparison}
	\vspace{-5mm}
\end{figure*}

\vspace{2mm}
\noindent{\bf Fast Density Reconstruction.}
NeVRF requires a density grid for each frame as a geometry proxy. 
%
%
%
%
To obtain density grids for $\mathbb{D}_{t_1}^{t_2}$, we first follow the learning procedure of DeVRF~\cite{liu2022devrf}, which speeds up the training by two orders of magnitude and avoids per-frame training. 
Then a 4D voxel deformation field $\mathbf{V}_{motion}$ together with a density grid $\mathbf{V}^{t_1}_{\sigma}$ and a feature grid in canonical space are obtained. 
Note that as the number of frames grows or with large motions, the rendering quality of DeVRF degenerates a lot, as demonstrated in Figure~\ref{fig:comparison}.  
However, the reconstructed geometry of the scene remains reasonable and can be refined in later stages.
%
We extract density grids from DeVRF and regard it as coarse proxy geometry and discard its appearance features. 
%
%
The obtained coarse density grids will be optimized further during NeVRF training.
%

\vspace{2mm}
\noindent{\bf Continual Neural Blending Learning.}
The blending network, denoted as $\Psi_{v}$, is designed to weight views based on their occlusion status. However, in dynamic scenes, occlusion conditions can change over time, necessitating the network to be updated with new incoming data.
%
%
%
Because of the sequential nature of the inputs, it is not feasible to sample from the entire sequence data assuming independence and identical distribution (i.i.d.) for joint batch training. Moreover, training a separate network for each frame individually is expensive and time-consuming. Therefore, we opt for a shared network for all frames. 
%
%
%
%
Nevertheless, updating the network on new data $\mathbb{D}_{t_1}^{t_2}$ is tricky. 
%
The over plasticity of neural networks~\cite{mirzadeh2020understanding} leads to degradation of the quality for previous frames when using the newly learned network parameters, known as {\it  catastrophic forgetting}~\cite{mccloskey1989catastrophic,french1999catastrophic}, according to the ”plasticity-stability dilemma”~\cite{mermillod2013stability}. 
Therefore, we adopt a continual learning technique, called {\it Experience Replay}~\cite{rolnick2019experience}, to handle this problem.

\begin{figure}[t]
	\begin{center}
		\includegraphics[width=1.0\linewidth]{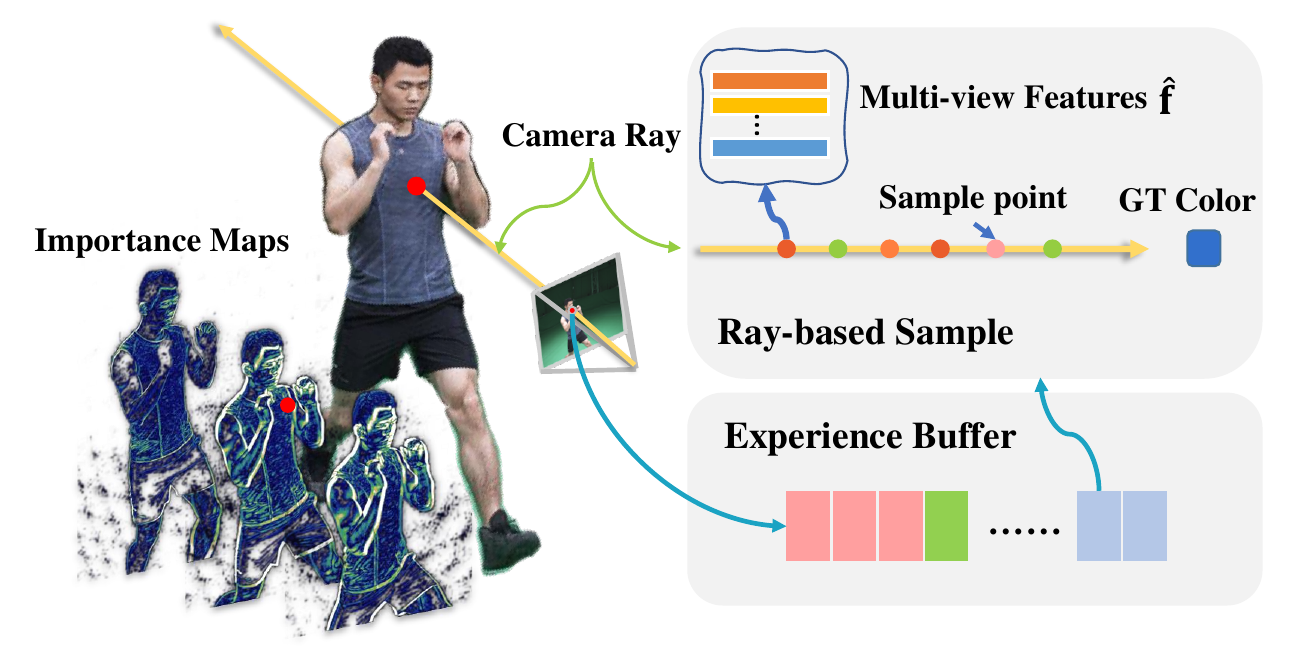}
	\end{center}
	\vspace{-5mm}
	\caption{ Illustration of the proposed ray-based replay strategy. The experience buffer contains samples sampled based on training errors (pink), temporal motions (green), and stochastic selection (blue). The importance maps visualize the pixels that have either large training error or large motion.}
	\label{fig:cl}
	\vspace{-4mm}
\end{figure}

Replay-based methods maintain an experience buffer $\mathcal{Q}^t$ that stores samples as the memory of previous knowledge. 
%
%
%
These samples remind the network how to process previous frames and help to retain this capability when training on sequential data.
Storing images in the buffer is inefficient because not all rays, especially those from background and textureless unoccluded regions, contribute to network learning. Therefore, we convert the image-based samples into ray-based ones for better efficiency.
%
%
%
As illustrated in Figure~\ref{fig:cl}, a ray-based sample $\mathbf{q}$ consists of multi-view feature sets of a ray and the ground-truth color of its corresponding pixel, i.e., $\mathbf{q}= (\{\mathbf{\hat{f}}_{i,j}\}_{i=1,j=1}^{n_s,k}, \mathbf{c})$ with $n_s$ the number of sample points along the ray and $\mathbf{c}$ the corresponding pixel's color. 
$\mathcal{Q}^t$ stores ray samples since the training beginning. 
Having this experience buffer, we can train the network with a loss function $\mathcal{L}_{T}$ formulated as:
\begin{equation}
\begin{split}
\mathcal{L}_{T}= \sum_{v=1}^{N_c}\mathcal{L}(f(\Phi(\mathcal{D}^t, v),\mathbf{\theta}),\mathbf{c})+\sum_{\mathcal{Q}^{t-1}}\mathcal{L}(f(\mathbf{q},\mathbf{\theta}),\mathbf{c})
\end{split}
\label{eq:loss}
\end{equation}
where $\mathcal{L}$ is the photometric loss; $f(\cdot)$ represents the radiance blending function as described in Sec.~\ref{sec:mrb}, and $\theta$ is their current network parameters. $\Phi(\mathcal{D}^t, v)$ denotes the multi-view image features for rendering the $v$-th view in the current frame $t$.

We want to store ray-based samples that contain as much representative knowledge as possible, so that the networks can preserve previous knowledge with a small experience buffer size. 
To this end, NeVRF selects ray-based samples according to their importance. 
We consider rays are important if their training loss or color difference in two consecutive frames is large. 
More specifically, we split the experience buffer into three parts: error-based buffer $\mathcal{Q}^{t-1}_e$, motion-based buffer $\mathcal{Q}^{t-1}_m$, and stochastic buffer $\mathcal{Q}^{t-1}_r$ respectively. 
That is: $\mathcal{Q}^{t-1} = \mathcal{Q}^{t-1}_e \cup \mathcal{Q}^{t-1}_m \cup \mathcal{Q}^{t-1}_r. $
%
In error-based buffer $\mathcal{Q}^{t-1}_e$, we sort ray-based samples according to their training errors in descending order and keep only the top-ranked samples. 
%
%
We append rays in the training batch to the motion-based buffer $\mathcal{Q}^{t-1}_m$ if their color difference with the next frame is above a threshold $\tau$. 
The stochastic buffer $\mathcal{Q}^{t-1}_r$ stores samples randomly selected from the training batch to provide diversity to the experience buffer. 
$20\%$ of ray samples in a training batch come from the experience buffer.
%
%
When the motion-based or stochastic buffer exceeds capacity, we randomly discard samples from it. 
%
%

Our training scheme trains each frame of $\mathbb{D}_{t_1}^{t_2}$  in temporal order. 
To improve the training performance, we only conduct continual learning on the first frame of $\mathbb{D}_{t_1}^{t_2}$. 
We fix networks and only optimize the density field for the rest of the frames with fewer iterations.
%

%
\begin{figure}[t]
	\begin{center}
		\includegraphics[width=1.0\linewidth]{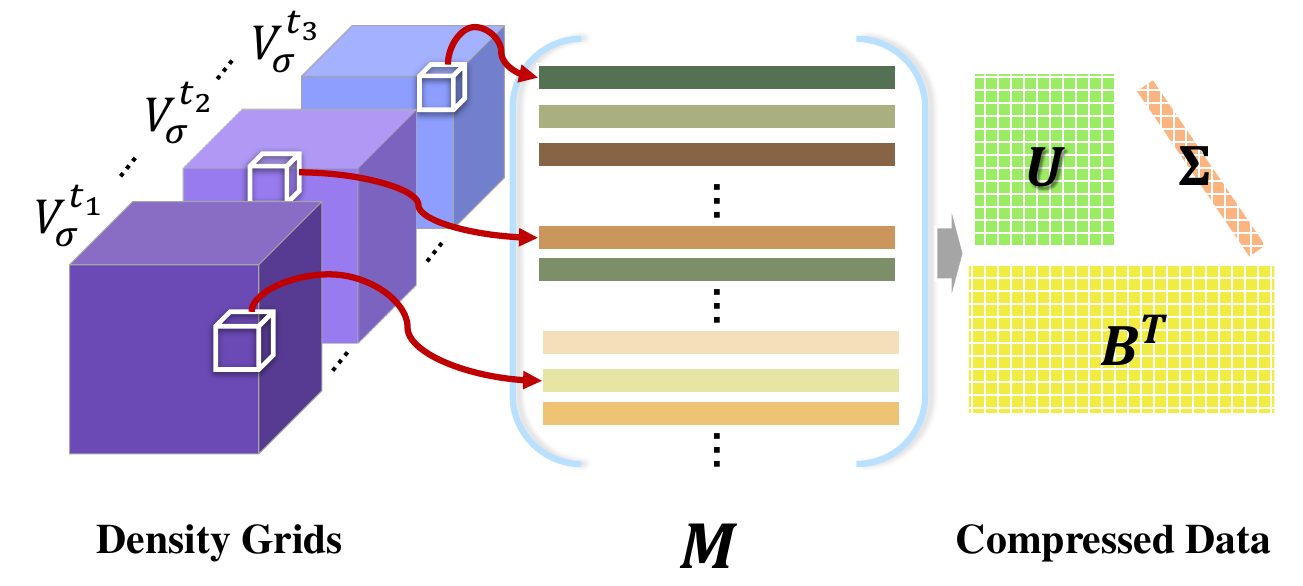}
	\end{center}
	\vspace{-0.5cm}
	\caption{The demonstration of the proposed density grid compression approach.}
	\label{fig:svd}
	\vspace{-5mm}
\end{figure}


\vspace{2mm}
\noindent{\bf Density Fields Compression.}
%
%
%
The density grids exhibit local structural similarity across frames in a sequence.
We set out to leverage this kind of redundancy on geometry to compress density grid sequences, as demonstrated in Figure~\ref{fig:svd}.

\begin{figure*}[t]
	\begin{center}
		\includegraphics[width=1.0\linewidth]{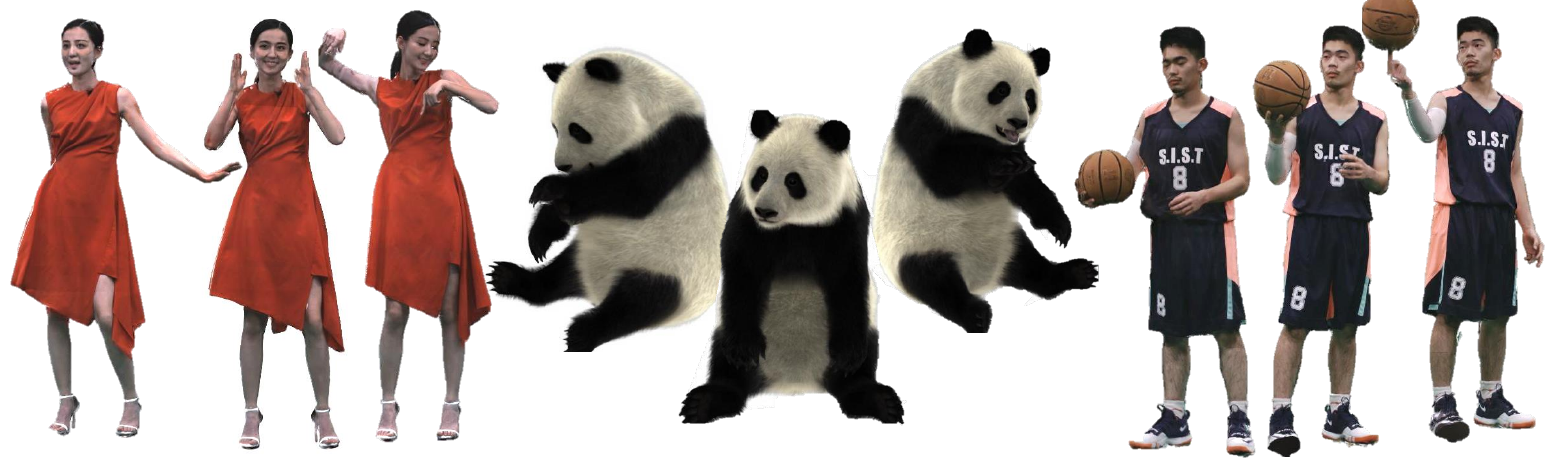}
	\end{center}
	\vspace{-4mm}
	\caption{Gallery of our example results. Our neural pipeline enables efficient training and photo-realistic rendering for dynamic scenes. Note that the results are all rendered at novel viewpoints}
	\label{fig:gallery}
 \vspace{-4mm}
\end{figure*}

Before training begins, we first align the 3D bounding boxes of density grids across all frames according to the camera positions. 
%
%
This alignment of density grids is beneficial for our compression algorithm, as it ensures uniform sizes and locations.
%
We compress the density grids of $\mathbb{D}_{t_1}^{t_2}$ simultaneously. 
These grids are divided into voxels of size $8\times8\times 8$ which are then flattened into vectors. 
This process results in a data matrix $\mathbf{M}\in\mathbb{R}^{N_v \times 512}$, where each row corresponds to a voxel from one frame, and $N_v$ is the total number of voxels over all frames considered.  
Subsequently, we use singular value decomposition~(SVD) to decompose this data matrix.
%
We retain only the top $\eta=20\%$ of the singular values to discard insignificant signals, resulting in the compressed matrix $\mathbf{M}=\mathbf{U}\mathbf{\Sigma}\mathbf{B}^T$, where $\mathbf{U}\in \mathbb{R}^{N_v\times k}$, $\mathbf{B}^T\in \mathbb{R}^{k\times 512}$, and $k=\eta\times N_v$. 
%
%
We consider a voxel to be empty if the sum of its raw density values is below zero.  
To further compress the data, we remove the rows corresponding to the empty voxels in $\mathbf{U}$. 

\subsection{Implementation Details}
%
%
%
%
%
%
%
Each $\mathbb{D}$ contains $20$ frames.
%
The multi-view videos are compressed by H265 codec with 1Mbps bitrate for each view. 
%



\section{Experiments}

In this section, we compare and evaluate the proposed NeVRF on a variety of dynamic scenes. 
Specifically, we first use the NHR dataset~\cite{wu2020multi} 
%
which contains large motions and topological changes.
We also test methods on Dynamic Furry Animals (DFA) dataset~\cite{luo2022artemis}, which is a non-human dataset. 
%
%

We set the target rendering resolution to $960\times720$ for NHR dataset and $960\times540$ for DFA dataset. 
We split the views into a training set and a test set, with the test set consisting of four views for each scene. 
%
%
%
%
We run our experiments on a single NVIDIA V100 GPU. 
Figure~\ref{fig:gallery} demonstrates some rendering results of our NeVRF. 
%

\begin{figure*}[t]
	\begin{center}
		\includegraphics[width=1\linewidth]{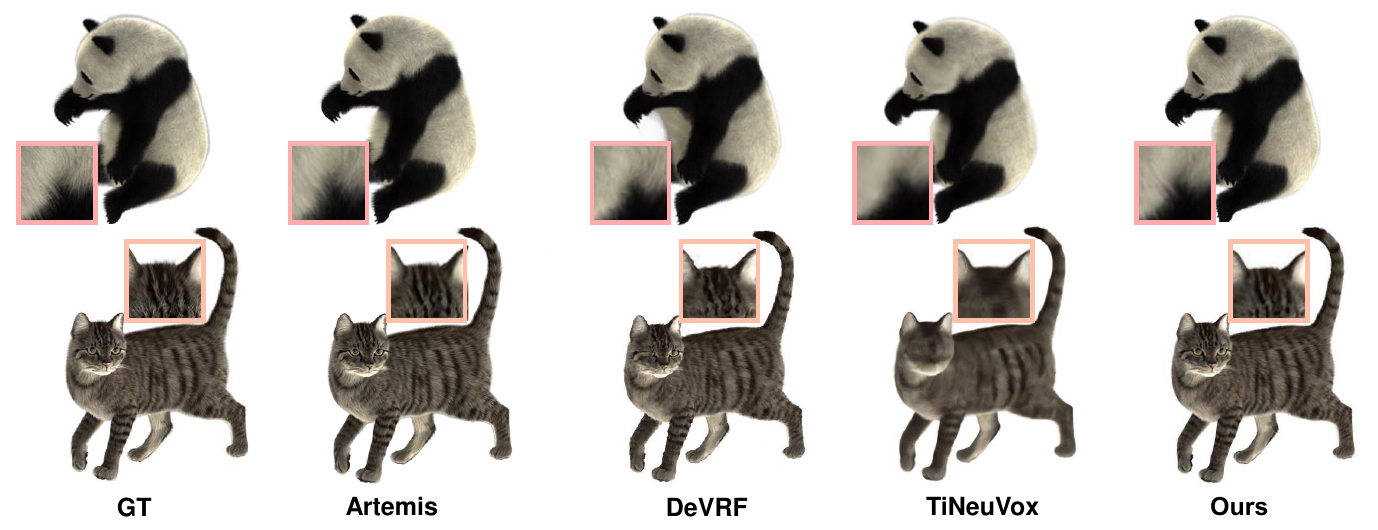}
	\end{center}
	\vspace{-4mm}
	\caption{Qualitative comparisons on DFA~\cite{luo2022artemis} dataset. DeVRF and TiNeuVox obtain distorted or blurred results on large motion parts.}
	\label{fig:comparison_animal}
	\vspace{-4mm}
\end{figure*}

\begin{figure}[t]
	\begin{center}
		\includegraphics[width=1.0\linewidth]{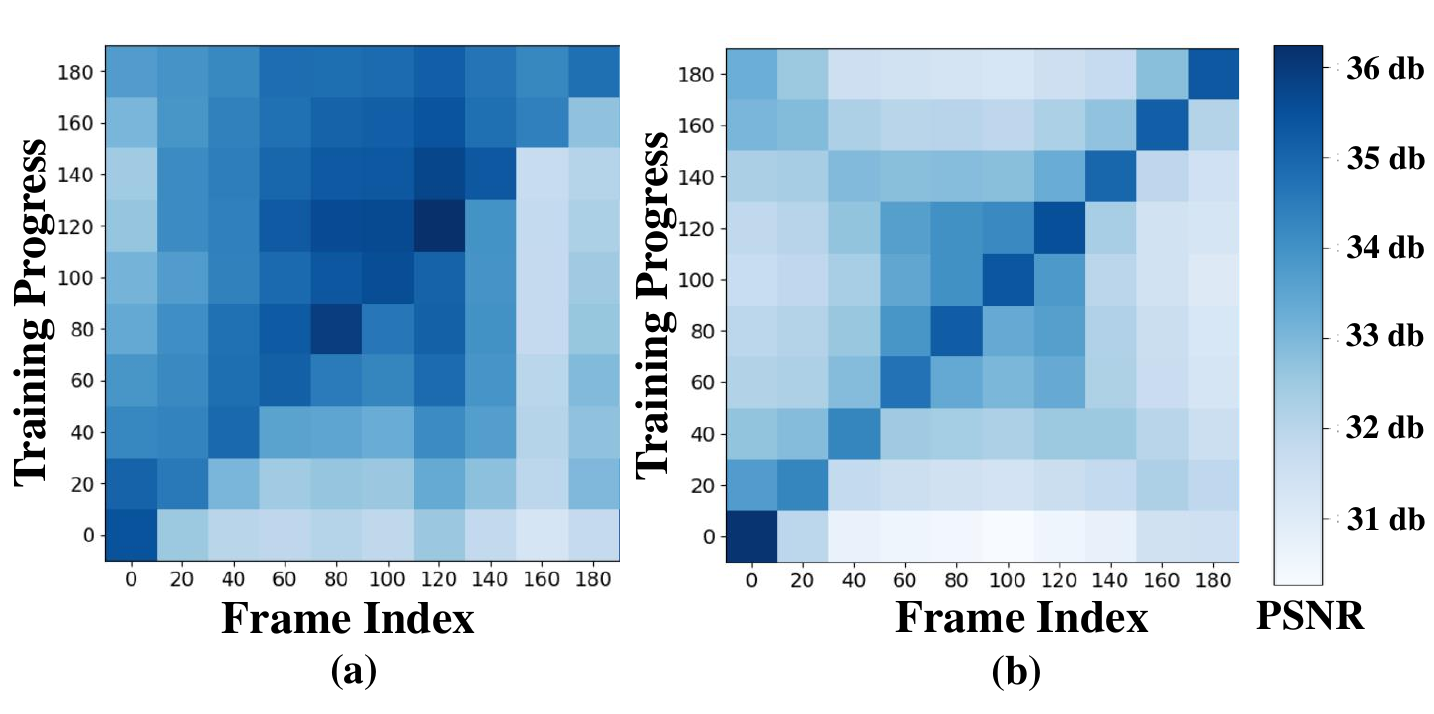}
	\end{center}
	\vspace{-0.5cm}
	\caption{The illustration of the intermediate results. The value of training progress indicates the number of frames after which the model used for evaluation was trained. (a) results with continual learning;(b) results without continual learning.}
	\label{fig:exp_replay}
	\vspace{-3mm}
\end{figure}

\subsection{Comparison}

\textbf{Comparison on NHR Dataset.} 
We compare NeVRF with the state-of-the art dynamic NeRF methods, DeVRF~\cite{liu2022devrf} and TiNeuvox~\cite{fang2022fast}, on NHR dataset. 
%
%
%
The majority of sequences from NHR are approximately 200 frames in length. 
Loading all this data into GPU memory is impractical, especially for sequences of longer duration. 
Consequently, for DeVRF and TiNeuvox training, we restrict our analysis to the initial $100$ frames of each sequence.
%
%
Despite the fact that DeVRF and TiNeuVox support sparse view inputs, we still use all training views in the experiments for fair comparisons. 
%
%
%
We also compare NeVRF with DVGO~\cite{sun2022direct}. 
We use DVGO to train each frame individually.
%
The quantitative results are shown in Table~\ref{tab:comparison}. 
Figure~\ref{fig:comparison} demonstrates the qualitative results.

As the results demonstrate, DVGO maintains high-quality rendering with fast rendering speed, although it requires large storage per frame. 
%
DeVRF greatly accelerates the model training. 
TiNeuVox, with its more implicit representation through MLPs, offers compact storage but incurs high costs in training and rendering time. 
%
Both DeVRF and TiNeuVox exhibit limitations when dealing with large motions and topological changes.
%
%
TiNeuVox gets bad rendering performance due to the limited network capacity, while DeVRF struggles to track motions due to the fixed topology of the canonical space. 
%
%
As a result, both models exhibit serious distortions in appearance and reconstructed geometry. 
%
%

%
In contrast, our method, which blends appearance directly from multi-view inputs, preserves high-frequency details.
Paired with video codecs and our density compression algorithm, NeVRF achieves a relatively small storage size.

%
%

\textbf{Comparison on DFA Dataset.} 
We also compare NeVRF with DeVRF~\cite{liu2022devrf}, TiNeuvox~\cite{fang2022fast} and Artemis~\cite{luo2022artemis} on a non-human dataset. 
%
%
We adopt its dynamic neural rendering comparison setup and test methods on 'Panda', 'Cat', 'Dog', and 'Lion' scenes. 
Again, we limit our comparisons to the first $100$ frames of each sequence or all frames if the sequence length is shorter than $100$. 
The quantitative results are shown in Table~\ref{tab:comparison_DFA}. 
Figure~\ref{fig:comparison_animal} demonstrates the qualitative results.

\begin{table}[]
\small
\setlength{\tabcolsep}{1.4mm}{
\begin{tabular}{l|cccccc}
\hline
                                 &$\uparrow$PSNR& $\uparrow$SSIM & $\downarrow$LPIPS & \makecell[c]{$\downarrow$Size\\(MB)} & \makecell[c]{$\downarrow$T.T.\\(mins)}& \makecell[c]{$\downarrow$R.T\\(secs)}\\ \hline
DVGO                             &  30.24     &  0.968    &  0.042 &   624      &  6              &  0.3              \\ \hline
DeVRF                            & 25.62     & 0.939      &  0.074  &   60.3      &    0.25           &   1.6             \\ 
TiNeuvox &  26.08     &  0.942   &  0.075     &    0.7     &     1.3      &   8.5             \\ \hline
ours                             &  32.15    &  0.976    &  0.034     &   1.32      &  3              &  1.9 \\ \hline              
\end{tabular}
}
\caption{The quantitative results of comparisons on NHR dataset. T.T. is the averaged per-frame training time, and R.T. is the averaged per-frame rendering time. 'Size' is the averaged per-frame storage size including compressed density, videos, and networks.
}
\label{tab:comparison}

\vspace{-3mm}
\end{table}

\begin{table}[]
\small
\setlength{\tabcolsep}{1.5mm}{
\begin{tabular}{lccccc}
\hline
Method                          &                    & DeVRF & TiNeuVox & Artemis & Ours \\ \hline
\multirow{3}{*}{\textbf{Panda}} & $\uparrow$PSNR    & 32.88 &  36.51   &   33.63 &  41.19    \\
                                & $\uparrow$SSIM    & 0.973 &  0.963   &    0.985&  0.985    \\
                                & $\downarrow$LPIPS & 0.028 &  0.058   &   0.031 &  0.014    \\ \hline
\multirow{3}{*}{\textbf{Cat}}   & $\uparrow$PSNR    & 32.82 &  31.354  &   37.54 &  40.29    \\
                                & $\uparrow$SSIM    & 0.969 &  0.948  &   0.989 &  0.987    \\
                                & $\downarrow$LPIPS & 0.026 &  0.051  &   0.012 &  0.010    \\ \hline
\multirow{3}{*}{\textbf{Dog}}   & $\uparrow$PSNR    &  30.37&   34.58  &  38.95  &  40.85\\
                                & $\uparrow$SSIM    &  0.964&   0.966  &  0.989  &  0.989\\
                                & $\downarrow$LPIPS &  0.046&   0.036  &  0.022  &  0.011 \\ \hline
\multirow{3}{*}{\textbf{Lion}}  & $\uparrow$PSNR    & 29.09 &   32.78  &  33.09  &  38.52 \\
                                & $\uparrow$SSIM    & 0.936 &   0.930  &  0.966  &  0.972\\
                                & $\downarrow$LPIPS & 0.058 &   0.084  &  0.035  &  0.021     \\ \hline
\end{tabular}
}
\caption{The quantitative results of comparisons on DFA dataset. }
\label{tab:comparison_DFA}
\vspace{-5mm}
\end{table}

\subsection{Ablation Study}
In this subsection, we evaluate the performance of our approach by comparing different training schemes and compression parameters. 
%

\textbf{Continual Learning Strategy.} 
To evaluate the effectiveness of the proposed continual neural blending learning scheme, we save all trained network models after training each group of data in the sequential input.
Using these intermediate models, we can evaluate the performance of previous and future frames at various stages of the training process to determine if the replay strategy helps mitigate the issue of {\it catastrophic forgetting}. 
%
%
We evaluate our approach using all frames of the "Sport1" scene and present the per-frame quality with and without continual learning in Figure~\ref{fig:exp_replay}. 
%
%
%
The upper left corner of the continual learning results is darker(higher db values) than the other areas, indicating that the network still performs well on previous frames.
%
%

\begin{figure}[t]
	\begin{center}
		\includegraphics[width=1.0\linewidth]{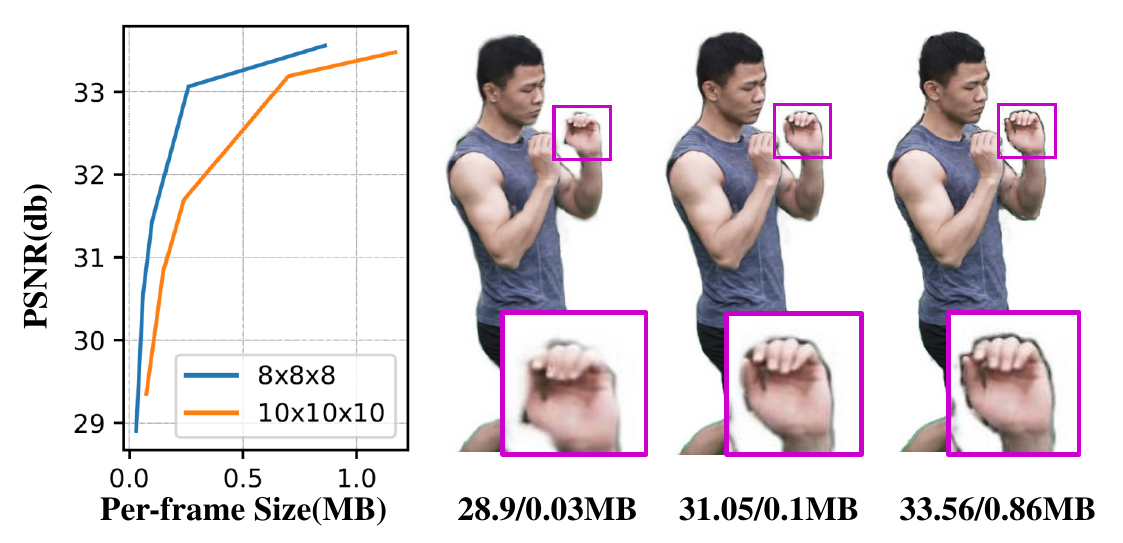}
	\end{center}
	\vspace{-0.5cm}
	\caption{ Demonstration of the compression results. The comparison of different voxel sizes is on the left. On the right are the qualitative results of different compression rates under the $8\times8\times8$ setting. Values under the results are the averaged PSNR and the per-frame storage size of only the density grids.}
	
	\label{fig:svd_res}
	\vspace{-5mm}
\end{figure}

\begin{figure}[t]
	\begin{center}
		\includegraphics[width=1.0\linewidth]{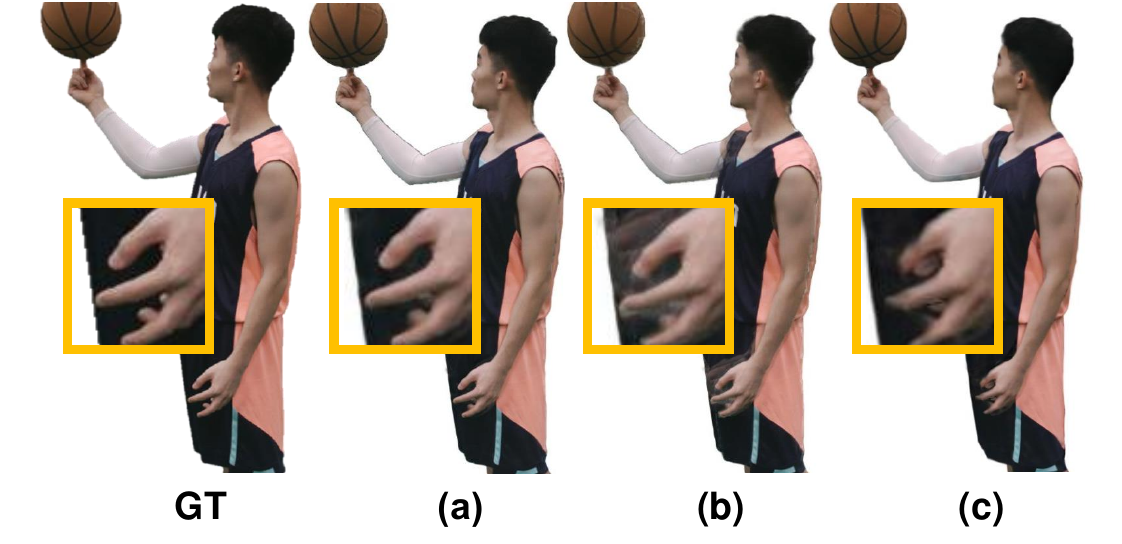}
	\end{center}
	\vspace{-0.5cm}
	\caption{Ablation Study. (a) Our full model; (b) Without the blending network; (c) Without density optimization. }
	
	\label{fig:visibility_net}
\end{figure}

\textbf{The Blending Network and Density Optimization.} 
%
We enable and disable the blending network and density optimization separately when training NeVRF models and test after that. 
As we can see in Figure~\ref{fig:visibility_net}, NeVRF without the blending network causes ghosting effects. 
Our end-to-end pipeline can refine the density and get better results.

\begin{figure}[t]
	\begin{center}
		\includegraphics[width=0.9\linewidth]{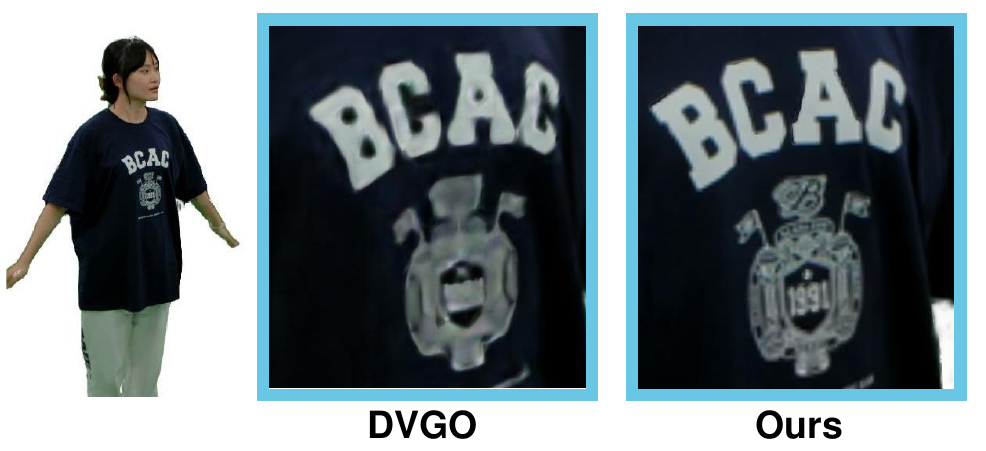}
	\end{center}
	\vspace{-0.5cm}
	\caption{Moving camera closer to the object. DVGO produces blurry textures while NeVRF is capable of preserving details.}
	
	\label{fig:zoomin}
\end{figure}

\textbf{Image Zoom-In.} 
The image-based rendering scheme accurately projects appearance from reference views onto the 3D space, enabling high-frequency details.
In this experiment, we move the target camera progressively closer to the object. 
Figure~\ref{fig:zoomin} demonstrates the qualitative results. 

\textbf{Density Compression.} We test two compression configurations as demonstrated in Figure~\ref{fig:svd_res}. 
Using a smaller voxel size of $8\times8\times8$ can compress the density fields more efficiently. 
The qualitative comparison in this figure shows that even though the density field is compressed to the extreme, NeVRF can still produce plausible results due to our image-based neural blending approach. 
Note that NeVRF is able to achieve smaller storage size by adjusting $\eta$. 

\begin{figure}[t]
\vspace{-5mm}
	\begin{center}
		\includegraphics[width=1.0\linewidth]{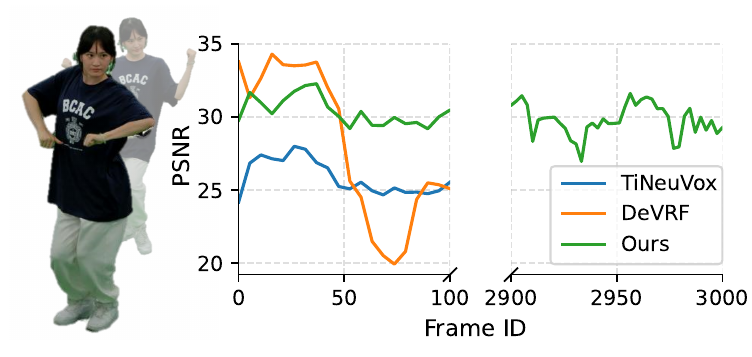}
	\end{center}
	\vspace{-0.5cm}
	\caption{TiNeuVox and DeVRF need access to all training data simultaneously and therefore can only cope with around 100 frames before running out of memory. In contrast, our method gets similar performance yet can keep processing new frames (results are shown for our model trained on a total of 3000 frames).}

	\label{fig:jywq}
	\vspace{-4mm}
\end{figure}

\textbf{Ultra Long Sequence.} 
We test methods on a sequence with 3000 frames and report the per-frame PSNR of different methods after entire training, as shown in Figure~\ref{fig:jywq}. 
As frames increases, DeVRF degenerates while TiNeuVox attains low PSNR and cannot handle the full length video as it cannot load all data into memory during training. 
NeVRF can still perform well at the end of the sequence.

\section{Limitation and Conclusion}
%
%
%
%
Similar to previous IBR methods~\cite{buehler2001unstructured, gortler1996lumigraph}, NeVRF also relies on dense reference views to minimize angular deviation for good rendering quality. 
However, the multi-view radiance blending is not well-suited for settings with very sparse views. And there is still room for improvement in speed.

%
%
%
%

We introduce a new Neural video-based Radiance Field (NeVRF) technique that efficiently models long-duration dynamic scenes in a compact form. Our radiance blending approach effectively infers radiance from multi-view videos, taking into account camera visibility. The innovative sequential training strategy with continual learning overcomes network {\it  `catastrophic forgetting`}, enabling long-duration sequence training with low memory footprint. NeVRF compresses dynamic scenes into a small size while maintaining high-quality rendering. Extensive tests show our method's ability to reconstruct and render long-duration sequences effectively.

%

\textbf{Acknowledgments.} 
This work was supported by the European Research Council (ERC) under the European Union's Horizon 2020 research and innovation programme (grant agreement n° 101021347) and the Flanders AI Research program.

{
    \small
    \bibliographystyle{ieeenat_fullname}
    \bibliography{main}

\begin{thebibliography}{54}
\providecommand{\natexlab}[1]{#1}
\providecommand{\url}[1]{\texttt{#1}}
\expandafter\ifx\csname urlstyle\endcsname\relax
  \providecommand{\doi}[1]{doi: #1}\else
  \providecommand{\doi}{doi: \begingroup \urlstyle{rm}\Url}\fi

\bibitem[Aliev et~al.(2020)Aliev, Sevastopolsky, Kolos, Ulyanov, and
  Lempitsky]{aliev2020neural}
Kara-Ali Aliev, Artem Sevastopolsky, Maria Kolos, Dmitry Ulyanov, and Victor
  Lempitsky.
\newblock Neural point-based graphics.
\newblock In \emph{European Conference on Computer Vision}, pages 696--712.
  Springer, 2020.

\bibitem[Buehler et~al.(2001)Buehler, Bosse, McMillan, Gortler, and
  Cohen]{buehler2001unstructured}
Chris Buehler, Michael Bosse, Leonard McMillan, Steven Gortler, and Michael
  Cohen.
\newblock Unstructured lumigraph rendering.
\newblock In \emph{Proceedings of the 28th annual conference on Computer
  graphics and interactive techniques}, pages 425--432, 2001.

\bibitem[Campbell et~al.(2008)Campbell, Vogiatzis, Hern{\'a}ndez, and
  Cipolla]{campbell2008using}
Neill~DF Campbell, George Vogiatzis, Carlos Hern{\'a}ndez, and Roberto Cipolla.
\newblock Using multiple hypotheses to improve depth-maps for multi-view
  stereo.
\newblock In \emph{European Conference on Computer Vision}, pages 766--779.
  Springer, 2008.

\bibitem[Casas et~al.(2015)Casas, Richardt, Collomosse, Theobalt, and
  Hilton]{casas20154d}
Dan Casas, Christian Richardt, John Collomosse, Christian Theobalt, and Adrian
  Hilton.
\newblock 4d model flow: Precomputed appearance alignment for real-time 4d
  video interpolation.
\newblock In \emph{Computer Graphics Forum}, pages 173--182. Wiley Online
  Library, 2015.

\bibitem[Chen et~al.(2022)Chen, Xu, Geiger, Yu, and Su]{chen2022tensorf}
Anpei Chen, Zexiang Xu, Andreas Geiger, Jingyi Yu, and Hao Su.
\newblock Tensorf: Tensorial radiance fields.
\newblock \emph{arXiv preprint arXiv:2203.09517}, 2022.

\bibitem[Daribo and Saito(2011)]{daribo2011novel}
Isma{\"e}l Daribo and Hideo Saito.
\newblock A novel inpainting-based layered depth video for 3dtv.
\newblock \emph{IEEE Transactions on Broadcasting}, 57\penalty0 (2):\penalty0
  533--541, 2011.

\bibitem[De~Lange et~al.(2021)De~Lange, Aljundi, Masana, Parisot, Jia,
  Leonardis, Slabaugh, and Tuytelaars]{de2021continual}
Matthias De~Lange, Rahaf Aljundi, Marc Masana, Sarah Parisot, Xu Jia,
  Ale{\v{s}} Leonardis, Gregory Slabaugh, and Tinne Tuytelaars.
\newblock A continual learning survey: Defying forgetting in classification
  tasks.
\newblock \emph{IEEE transactions on pattern analysis and machine
  intelligence}, 44\penalty0 (7):\penalty0 3366--3385, 2021.

\bibitem[Debevec et~al.(1996)Debevec, Taylor, and Malik]{debevec1996modeling}
Paul~Ernest Debevec, Camillo~J Taylor, and Jitendra Malik.
\newblock \emph{Modeling and rendering architecture from photographs}.
\newblock University of California, Berkeley, 1996.

\bibitem[Deng et~al.(2022)Deng, Liu, Zhu, and Ramanan]{deng2022depth}
Kangle Deng, Andrew Liu, Jun-Yan Zhu, and Deva Ramanan.
\newblock Depth-supervised nerf: Fewer views and faster training for free.
\newblock In \emph{Proceedings of the IEEE/CVF Conference on Computer Vision
  and Pattern Recognition}, pages 12882--12891, 2022.

\bibitem[Du et~al.(2018)Du, Chuang, Chang, Hoppe, and
  Varshney]{du2018montage4d}
Ruofei Du, Ming Chuang, Wayne Chang, Hugues Hoppe, and Amitabh Varshney.
\newblock Montage4d: interactive seamless fusion of multiview video textures.
\newblock In \emph{Proceedings of the ACM SIGGRAPH Symposium on Interactive 3D
  Graphics and Games}, pages 1--11, 2018.

\bibitem[Du et~al.(2021)Du, Zhang, Yu, Tenenbaum, and Wu]{du2021neural}
Yilun Du, Yinan Zhang, Hong-Xing Yu, Joshua~B Tenenbaum, and Jiajun Wu.
\newblock Neural radiance flow for 4d view synthesis and video processing.
\newblock In \emph{2021 IEEE/CVF International Conference on Computer Vision
  (ICCV)}, pages 14304--14314. IEEE Computer Society, 2021.

\bibitem[Fang et~al.(2022)Fang, Yi, Wang, Xie, Zhang, Liu, Nie{\ss}ner, and
  Tian]{fang2022fast}
Jiemin Fang, Taoran Yi, Xinggang Wang, Lingxi Xie, Xiaopeng Zhang, Wenyu Liu,
  Matthias Nie{\ss}ner, and Qi Tian.
\newblock Fast dynamic radiance fields with time-aware neural voxels.
\newblock \emph{arXiv preprint arXiv:2205.15285}, 2022.

\bibitem[French(1999)]{french1999catastrophic}
Robert~M French.
\newblock Catastrophic forgetting in connectionist networks.
\newblock \emph{Trends in cognitive sciences}, 3\penalty0 (4):\penalty0
  128--135, 1999.

\bibitem[Garbin et~al.(2021)Garbin, Kowalski, Johnson, Shotton, and
  Valentin]{garbin2021fastnerf}
Stephan~J Garbin, Marek Kowalski, Matthew Johnson, Jamie Shotton, and Julien
  Valentin.
\newblock Fastnerf: High-fidelity neural rendering at 200fps.
\newblock In \emph{Proceedings of the IEEE/CVF International Conference on
  Computer Vision}, pages 14346--14355, 2021.

\bibitem[Gortler et~al.(1996)Gortler, Grzeszczuk, Szeliski, and
  Cohen]{gortler1996lumigraph}
Steven~J Gortler, Radek Grzeszczuk, Richard Szeliski, and Michael~F Cohen.
\newblock The lumigraph.
\newblock In \emph{Siggraph}, pages 43--54, 1996.

\bibitem[Isele and Cosgun(2018)]{isele2018selective}
David Isele and Akansel Cosgun.
\newblock Selective experience replay for lifelong learning.
\newblock In \emph{Proceedings of the AAAI Conference on Artificial
  Intelligence}, 2018.

\bibitem[Levoy and Hanrahan(1996)]{levoy1996light}
Marc Levoy and Pat Hanrahan.
\newblock Light field rendering.
\newblock In \emph{Proceedings of the 23rd annual conference on Computer
  graphics and interactive techniques}, pages 31--42, 1996.

\bibitem[Li et~al.(2022)Li, Slavcheva, Zollhoefer, Green, Lassner, Kim,
  Schmidt, Lovegrove, Goesele, Newcombe, et~al.]{li2022neural}
Tianye Li, Mira Slavcheva, Michael Zollhoefer, Simon Green, Christoph Lassner,
  Changil Kim, Tanner Schmidt, Steven Lovegrove, Michael Goesele, Richard
  Newcombe, et~al.
\newblock Neural 3d video synthesis from multi-view video.
\newblock In \emph{Proceedings of the IEEE/CVF Conference on Computer Vision
  and Pattern Recognition}, pages 5521--5531, 2022.

\bibitem[Lin et~al.(2022)Lin, Peng, Xu, Yan, Shuai, Bao, and
  Zhou]{lin2022efficient}
Haotong Lin, Sida Peng, Zhen Xu, Yunzhi Yan, Qing Shuai, Hujun Bao, and Xiaowei
  Zhou.
\newblock Efficient neural radiance fields for interactive free-viewpoint
  video.
\newblock In \emph{SIGGRAPH Asia 2022 Conference Papers}, pages 1--9, 2022.

\bibitem[Liu et~al.(2022)Liu, Cao, Mao, Zhang, Zhang, Keppo, Shan, Qie, and
  Shou]{liu2022devrf}
Jia-Wei Liu, Yan-Pei Cao, Weijia Mao, Wenqiao Zhang, David~Junhao Zhang, Jussi
  Keppo, Ying Shan, Xiaohu Qie, and Mike~Zheng Shou.
\newblock Devrf: Fast deformable voxel radiance fields for dynamic scenes.
\newblock \emph{arXiv preprint arXiv:2205.15723}, 2022.

\bibitem[Liu et~al.(2019)Liu, Xu, Zollhoefer, Kim, Bernard, Habermann, Wang,
  and Theobalt]{liu2019neural}
Lingjie Liu, Weipeng Xu, Michael Zollhoefer, Hyeongwoo Kim, Florian Bernard,
  Marc Habermann, Wenping Wang, and Christian Theobalt.
\newblock Neural rendering and reenactment of human actor videos.
\newblock \emph{ACM Transactions on Graphics (TOG)}, 38\penalty0 (5):\penalty0
  1--14, 2019.

\bibitem[Liu et~al.(2020)Liu, Gu, Zaw~Lin, Chua, and Theobalt]{liu2020neural}
Lingjie Liu, Jiatao Gu, Kyaw Zaw~Lin, Tat-Seng Chua, and Christian Theobalt.
\newblock Neural sparse voxel fields.
\newblock \emph{Advances in Neural Information Processing Systems},
  33:\penalty0 15651--15663, 2020.

\bibitem[Lombardi et~al.(2019)Lombardi, Simon, Saragih, Schwartz, Lehrmann, and
  Sheikh]{lombardi2019neural}
Stephen Lombardi, Tomas Simon, Jason Saragih, Gabriel Schwartz, Andreas
  Lehrmann, and Yaser Sheikh.
\newblock Neural volumes: Learning dynamic renderable volumes from images.
\newblock \emph{arXiv preprint arXiv:1906.07751}, 2019.

\bibitem[Luo et~al.(2022)Luo, Xu, Jiang, Zhou, Qiu, Zhang, Yang, Xu, and
  Yu]{luo2022artemis}
Haimin Luo, Teng Xu, Yuheng Jiang, Chenglin Zhou, QIwei Qiu, Yingliang Zhang,
  Wei Yang, Lan Xu, and Jingyi Yu.
\newblock Artemis: articulated neural pets with appearance and motion
  synthesis.
\newblock \emph{arXiv preprint arXiv:2202.05628}, 2022.

\bibitem[Matusik et~al.(2000)Matusik, Buehler, Raskar, Gortler, and
  McMillan]{matusik2000image}
Wojciech Matusik, Chris Buehler, Ramesh Raskar, Steven~J Gortler, and Leonard
  McMillan.
\newblock Image-based visual hulls.
\newblock In \emph{Proceedings of the 27th annual conference on Computer
  graphics and interactive techniques}, pages 369--374. ACM
  Press/Addison-Wesley Publishing Co., 2000.

\bibitem[Matusik et~al.(2002)Matusik, Pfister, Ngan, Beardsley, Ziegler, and
  Mcmillan]{Matusik2002Image}
Wojciech Matusik, Hanspeter Pfister, Addy Ngan, Paul Beardsley, Remo Ziegler,
  and Leonard Mcmillan.
\newblock Image-based 3d photography using opacity hulls.
\newblock \emph{Acm Transactions on Graphics}, 21\penalty0 (3):\penalty0
  427--437, 2002.

\bibitem[McCloskey and Cohen(1989)]{mccloskey1989catastrophic}
Michael McCloskey and Neal~J Cohen.
\newblock Catastrophic interference in connectionist networks: The sequential
  learning problem.
\newblock In \emph{Psychology of learning and motivation}, pages 109--165.
  Elsevier, 1989.

\bibitem[Mermillod et~al.(2013)Mermillod, Bugaiska, and
  Bonin]{mermillod2013stability}
Martial Mermillod, Aur{\'e}lia Bugaiska, and Patrick Bonin.
\newblock The stability-plasticity dilemma: Investigating the continuum from
  catastrophic forgetting to age-limited learning effects, 2013.

\bibitem[Mescheder et~al.(2019)Mescheder, Oechsle, Niemeyer, Nowozin, and
  Geiger]{mescheder2019occupancy}
Lars Mescheder, Michael Oechsle, Michael Niemeyer, Sebastian Nowozin, and
  Andreas Geiger.
\newblock Occupancy networks: Learning 3d reconstruction in function space.
\newblock In \emph{Proceedings of the IEEE/CVF conference on computer vision
  and pattern recognition}, pages 4460--4470, 2019.

\bibitem[Mildenhall et~al.(2021)Mildenhall, Srinivasan, Tancik, Barron,
  Ramamoorthi, and Ng]{mildenhall2021nerf}
Ben Mildenhall, Pratul~P Srinivasan, Matthew Tancik, Jonathan~T Barron, Ravi
  Ramamoorthi, and Ren Ng.
\newblock Nerf: Representing scenes as neural radiance fields for view
  synthesis.
\newblock \emph{Communications of the ACM}, 65\penalty0 (1):\penalty0 99--106,
  2021.

\bibitem[Mirzadeh et~al.(2020)Mirzadeh, Farajtabar, Pascanu, and
  Ghasemzadeh]{mirzadeh2020understanding}
Seyed~Iman Mirzadeh, Mehrdad Farajtabar, Razvan Pascanu, and Hassan
  Ghasemzadeh.
\newblock Understanding the role of training regimes in continual learning.
\newblock \emph{Advances in Neural Information Processing Systems},
  33:\penalty0 7308--7320, 2020.

\bibitem[M{\"u}ller et~al.(2022)M{\"u}ller, Evans, Schied, and
  Keller]{muller2022instant}
Thomas M{\"u}ller, Alex Evans, Christoph Schied, and Alexander Keller.
\newblock Instant neural graphics primitives with a multiresolution hash
  encoding.
\newblock \emph{arXiv preprint arXiv:2201.05989}, 2022.

\bibitem[Neff et~al.(2021)Neff, Stadlbauer, Parger, Kurz, Mueller, Chaitanya,
  Kaplanyan, and Steinberger]{neff2021donerf}
Thomas Neff, Pascal Stadlbauer, Mathias Parger, Andreas Kurz, Joerg~H Mueller,
  Chakravarty R~Alla Chaitanya, Anton Kaplanyan, and Markus Steinberger.
\newblock Donerf: Towards real-time rendering of compact neural radiance fields
  using depth oracle networks.
\newblock In \emph{Computer Graphics Forum}, pages 45--59. Wiley Online
  Library, 2021.

\bibitem[Park et~al.(2021)Park, Sinha, Barron, Bouaziz, Goldman, Seitz, and
  Martin-Brualla]{park2021nerfies}
Keunhong Park, Utkarsh Sinha, Jonathan~T Barron, Sofien Bouaziz, Dan~B Goldman,
  Steven~M Seitz, and Ricardo Martin-Brualla.
\newblock Nerfies: Deformable neural radiance fields.
\newblock In \emph{Proceedings of the IEEE/CVF International Conference on
  Computer Vision}, pages 5865--5874, 2021.

\bibitem[Peng et~al.(2021)Peng, Zhang, Xu, Wang, Shuai, Bao, and
  Zhou]{peng2021neural}
Sida Peng, Yuanqing Zhang, Yinghao Xu, Qianqian Wang, Qing Shuai, Hujun Bao,
  and Xiaowei Zhou.
\newblock Neural body: Implicit neural representations with structured latent
  codes for novel view synthesis of dynamic humans.
\newblock In \emph{Proceedings of the IEEE/CVF Conference on Computer Vision
  and Pattern Recognition}, pages 9054--9063, 2021.

\bibitem[Penner and Zhang(2017)]{penner2017soft}
Eric Penner and Li Zhang.
\newblock Soft 3d reconstruction for view synthesis.
\newblock \emph{ACM Transactions on Graphics (TOG)}, 36\penalty0 (6):\penalty0
  1--11, 2017.

\bibitem[Pumarola et~al.(2021)Pumarola, Corona, Pons-Moll, and
  Moreno-Noguer]{pumarola2021d}
Albert Pumarola, Enric Corona, Gerard Pons-Moll, and Francesc Moreno-Noguer.
\newblock D-nerf: Neural radiance fields for dynamic scenes.
\newblock In \emph{Proceedings of the IEEE/CVF Conference on Computer Vision
  and Pattern Recognition}, pages 10318--10327, 2021.

\bibitem[Rebuffi et~al.(2017)Rebuffi, Kolesnikov, Sperl, and
  Lampert]{rebuffi2017icarl}
Sylvestre-Alvise Rebuffi, Alexander Kolesnikov, Georg Sperl, and Christoph~H
  Lampert.
\newblock icarl: Incremental classifier and representation learning.
\newblock In \emph{Proceedings of the IEEE conference on Computer Vision and
  Pattern Recognition}, pages 2001--2010, 2017.

\bibitem[Reiser et~al.(2021)Reiser, Peng, Liao, and Geiger]{reiser2021kilonerf}
Christian Reiser, Songyou Peng, Yiyi Liao, and Andreas Geiger.
\newblock Kilonerf: Speeding up neural radiance fields with thousands of tiny
  mlps.
\newblock In \emph{Proceedings of the IEEE/CVF International Conference on
  Computer Vision}, pages 14335--14345, 2021.

\bibitem[Riegler and Koltun(2020)]{riegler2020free}
Gernot Riegler and Vladlen Koltun.
\newblock Free view synthesis.
\newblock In \emph{European Conference on Computer Vision}, pages 623--640.
  Springer, 2020.

\bibitem[Rolnick et~al.(2019)Rolnick, Ahuja, Schwarz, Lillicrap, and
  Wayne]{rolnick2019experience}
David Rolnick, Arun Ahuja, Jonathan Schwarz, Timothy Lillicrap, and Gregory
  Wayne.
\newblock Experience replay for continual learning.
\newblock \emph{Advances in Neural Information Processing Systems}, 32, 2019.

\bibitem[Saito et~al.(2019)Saito, Huang, Natsume, Morishima, Kanazawa, and
  Li]{saito2019pifu}
Shunsuke Saito, Zeng Huang, Ryota Natsume, Shigeo Morishima, Angjoo Kanazawa,
  and Hao Li.
\newblock Pifu: Pixel-aligned implicit function for high-resolution clothed
  human digitization.
\newblock In \emph{Proceedings of the IEEE/CVF International Conference on
  Computer Vision}, pages 2304--2314, 2019.

\bibitem[Shade et~al.(1998)Shade, Gortler, He, and Szeliski]{shade1998layered}
Jonathan Shade, Steven Gortler, Li-wei He, and Richard Szeliski.
\newblock Layered depth images.
\newblock In \emph{Proceedings of the 25th annual conference on Computer
  graphics and interactive techniques}, pages 231--242, 1998.

\bibitem[Sun et~al.(2022)Sun, Sun, and Chen]{sun2022direct}
Cheng Sun, Min Sun, and Hwann-Tzong Chen.
\newblock Direct voxel grid optimization: Super-fast convergence for radiance
  fields reconstruction.
\newblock In \emph{Proceedings of the IEEE/CVF Conference on Computer Vision
  and Pattern Recognition}, pages 5459--5469, 2022.

\bibitem[Suo et~al.(2021)Suo, Jiang, Lin, Zhang, Wu, Guo, and
  Xu]{suo2021neuralhumanfvv}
Xin Suo, Yuheng Jiang, Pei Lin, Yingliang Zhang, Minye Wu, Kaiwen Guo, and Lan
  Xu.
\newblock Neuralhumanfvv: Real-time neural volumetric human performance
  rendering using rgb cameras.
\newblock In \emph{Proceedings of the IEEE/CVF conference on computer vision
  and pattern recognition}, pages 6226--6237, 2021.

\bibitem[Takikawa et~al.(2022)Takikawa, Evans, Tremblay, M{\"u}ller, McGuire,
  Jacobson, and Fidler]{takikawa2022variable}
Towaki Takikawa, Alex Evans, Jonathan Tremblay, Thomas M{\"u}ller, Morgan
  McGuire, Alec Jacobson, and Sanja Fidler.
\newblock Variable bitrate neural fields.
\newblock In \emph{ACM SIGGRAPH 2022 Conference Proceedings}, pages 1--9, 2022.

\bibitem[Tola et~al.(2012)Tola, Strecha, and Fua]{tola2012efficient}
Engin Tola, Christoph Strecha, and Pascal Fua.
\newblock Efficient large-scale multi-view stereo for ultra high-resolution
  image sets.
\newblock \emph{Machine Vision and Applications}, 23\penalty0 (5):\penalty0
  903--920, 2012.

\bibitem[Wang et~al.(2022)Wang, Zhang, Liu, Zhao, Zhang, Zhang, Wu, Yu, and
  Xu]{wang2022fourier}
Liao Wang, Jiakai Zhang, Xinhang Liu, Fuqiang Zhao, Yanshun Zhang, Yingliang
  Zhang, Minye Wu, Jingyi Yu, and Lan Xu.
\newblock Fourier plenoctrees for dynamic radiance field rendering in
  real-time.
\newblock In \emph{Proceedings of the IEEE/CVF Conference on Computer Vision
  and Pattern Recognition}, pages 13524--13534, 2022.

\bibitem[Wang et~al.(2021)Wang, Wang, Genova, Srinivasan, Zhou, Barron,
  Martin-Brualla, Snavely, and Funkhouser]{wang2021ibrnet}
Qianqian Wang, Zhicheng Wang, Kyle Genova, Pratul~P Srinivasan, Howard Zhou,
  Jonathan~T Barron, Ricardo Martin-Brualla, Noah Snavely, and Thomas
  Funkhouser.
\newblock Ibrnet: Learning multi-view image-based rendering.
\newblock In \emph{Proceedings of the IEEE/CVF Conference on Computer Vision
  and Pattern Recognition}, pages 4690--4699, 2021.

\bibitem[Wu et~al.(2020)Wu, Wang, Hu, and Yu]{wu2020multi}
Minye Wu, Yuehao Wang, Qiang Hu, and Jingyi Yu.
\newblock Multi-view neural human rendering.
\newblock In \emph{Proceedings of the IEEE/CVF Conference on Computer Vision
  and Pattern Recognition}, pages 1682--1691, 2020.

\bibitem[Xian et~al.(2021)Xian, Huang, Kopf, and Kim]{xian2021space}
Wenqi Xian, Jia-Bin Huang, Johannes Kopf, and Changil Kim.
\newblock Space-time neural irradiance fields for free-viewpoint video.
\newblock In \emph{Proceedings of the IEEE/CVF Conference on Computer Vision
  and Pattern Recognition}, pages 9421--9431, 2021.

\bibitem[Yu et~al.(2021)Yu, Li, Tancik, Li, Ng, and
  Kanazawa]{yu2021plenoctrees}
Alex Yu, Ruilong Li, Matthew Tancik, Hao Li, Ren Ng, and Angjoo Kanazawa.
\newblock Plenoctrees for real-time rendering of neural radiance fields.
\newblock In \emph{Proceedings of the IEEE/CVF International Conference on
  Computer Vision}, pages 5752--5761, 2021.

\bibitem[Zhang et~al.(2021)Zhang, Liu, Ye, Zhao, Zhang, Wu, Zhang, Xu, and
  Yu]{zhang2021editable}
Jiakai Zhang, Xinhang Liu, Xinyi Ye, Fuqiang Zhao, Yanshun Zhang, Minye Wu,
  Yingliang Zhang, Lan Xu, and Jingyi Yu.
\newblock Editable free-viewpoint video using a layered neural representation.
\newblock \emph{ACM Transactions on Graphics (TOG)}, 40\penalty0 (4):\penalty0
  1--18, 2021.

\bibitem[Zhao et~al.(2022)Zhao, Jiang, Yao, Zhang, Wang, Dai, Zhong, Zhang, Wu,
  Xu, et~al.]{zhao2022human}
Fuqiang Zhao, Yuheng Jiang, Kaixin Yao, Jiakai Zhang, Liao Wang, Haizhao Dai,
  Yuhui Zhong, Yingliang Zhang, Minye Wu, Lan Xu, et~al.
\newblock Human performance modeling and rendering via neural animated mesh.
\newblock \emph{arXiv preprint arXiv:2209.08468}, 2022.

\end{thebibliography}
}

\end{document}